\documentclass{mva_style}
\usepackage{graphicx}

\finalcopy 

\newcommand{\FIG}[3]{
\begin{minipage}[b]{#1cm}
\begin{center}
\includegraphics[width=#1cm]{#2}\\
{\scriptsize #3}
\end{center}
\end{minipage}
}

\begin{document}

\newcommand{\tabA}{
\begin{table}[t]
\centering
\caption{Evaluation of training results for domain A}
  \begin{tabular}{|c|c|c|c|c|}\hline
		&A &B &C &D\\ \hline  \hline
	Proposed method  & 0.317 & 0.306 & 0.489 & 0.588 \\
	FPN   & 0.314 & 0.132 & 0.360 & 0.476 \\ \hline
  \end{tabular}
　　\label{Table:result_A}
\end{table}
}

\newcommand{\tabB}{
\begin{table}[t]
\centering
\caption{Evaluation of training results for domain B}
  \begin{tabular}{|c|c|c|c|c|}\hline
		&A &B &C &D\\ \hline  \hline
	Proposed method  & 0.334 & 0.521 & 0.411 & 0.445 \\
	FPN   & 0.179 & 0.397 & 0.143 & 0.088 \\ \hline
  \end{tabular}
　　\label{Table:result_B}
\end{table}
}

\newcommand{\tabAB}{
\begin{table}[t]
\centering
\caption{Evaluation of training on domains A and B}
  \begin{tabular}{|c|c|c|c|c|}\hline
		&A &B &C &D\\ \hline  \hline
	Proposed w/ training set A  & 0.317 & 0.306 & 0.489 & 0.588 \\
        Proposed w/ training set B & 0.317 & 0.306 & 0.489 & 0.588 \\ \hline
	FPN w/ training set A  & 0.314 & 0.132 & 0.360 & 0.476 \\ 
	FPN w/ training set B  & 0.314 & 0.132 & 0.360 & 0.476 \\ \hline
  \end{tabular}
\label{tab:AB}
\end{table}
}

\newcommand{\hh}[1]{\hspace*{-1mm}#1\hspace{-1mm}}

\renewcommand{\tabAB}{
\begin{table}[t]
{
\scriptsize
\centering
\caption{Average-precision performance [\%]}\label{tab:AB}
  \begin{tabular}{|l|r|r|r|r|r|r|r|r|}\hline
training set & \multicolumn{4}{|c|}{A} & \multicolumn{4}{|c|}{B} \\ \hline
test set & A & B &C &D & A &B &C &D \\ \hline  \hline
FPN & \hh{31.4} & \hh{13.2} & \hh{36.0} & \hh{47.6} & \hh{17.9} & \hh{39.7} & \hh{14.3} & \hh{8.8} \\ \hline
Proposed  & \hh{31.7} & \hh{30.6} & \hh{48.9} & \hh{58.8} & \hh{33.4} & \hh{52.1} & \hh{41.1} & \hh{44.5} \\ \hline
  \end{tabular}
}
\end{table}
}

\newcommand{\figA}{
  \begin{figure}[t]
    \FIG{8}{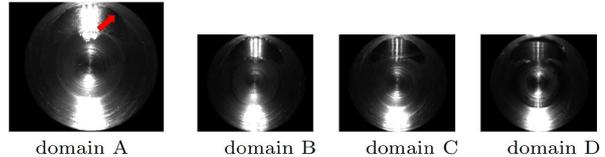}{}
  \caption{Visual burr detection in HPPPs.}\label{fig:rei1}
\end{figure}
}

\newcommand{\figB}{
\begin{figure}[t]
  \centering
  \FIG{8}{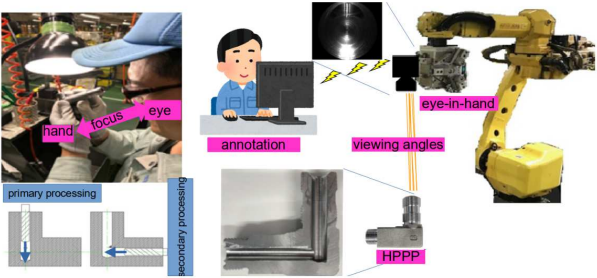}{}
\caption{%
From manual to automatic VDD.
}\label{fig:eye}
\end{figure}
}

\newcommand{\figC}{
\begin{figure}[t]
\FIG{8}{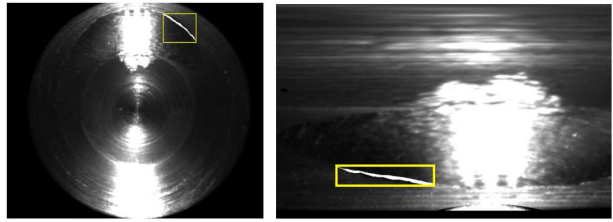}{}~\\
\caption{Ground-truth burrs and annotations.}\label{fig:conversion}
\end{figure}
}

\newcommand{\figD}{
\begin{figure}[t]
  \centering
\FIG{8}{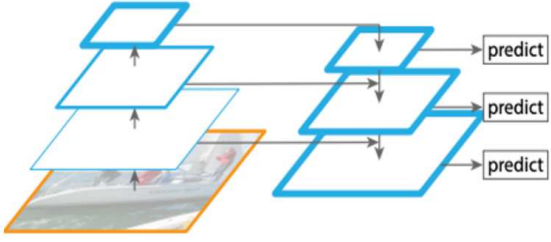}{}~\\
  \caption{NAS-searched object-detector.}\label{fig:D}
\label{fig:FPN}
\end{figure}
}

\newcommand{\figG}{
\begin{figure}[t]
\FIG{3}{test_valid_3.eps}{}~\\
  \caption{Evaluation result of training result by search data}
\label{fig:test_vald}
\end{figure}
}

\newcommand{\figH}{
\begin{figure}[t]
  \centering
  \scriptsize
\FIG{6}{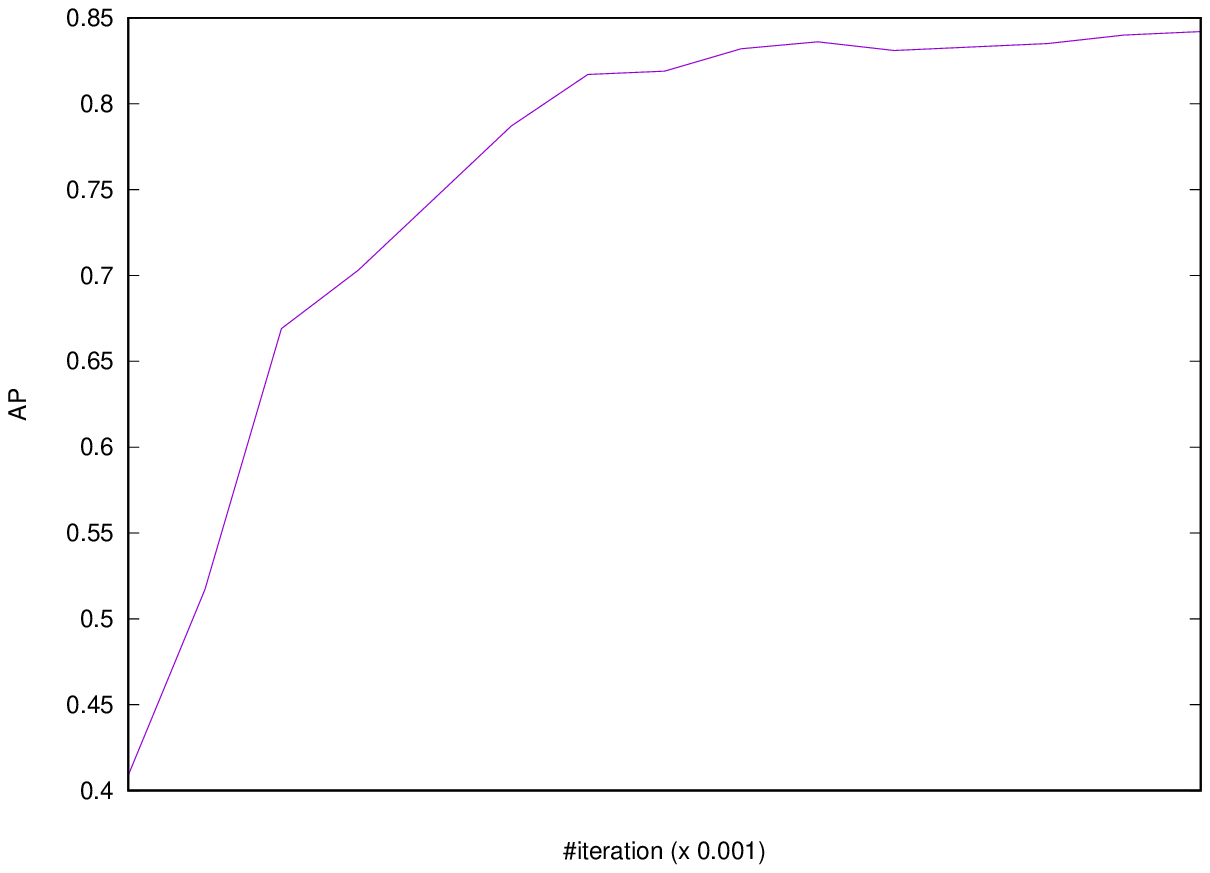}{}\vspace*{8mm}\\
a\\
\FIG{6}{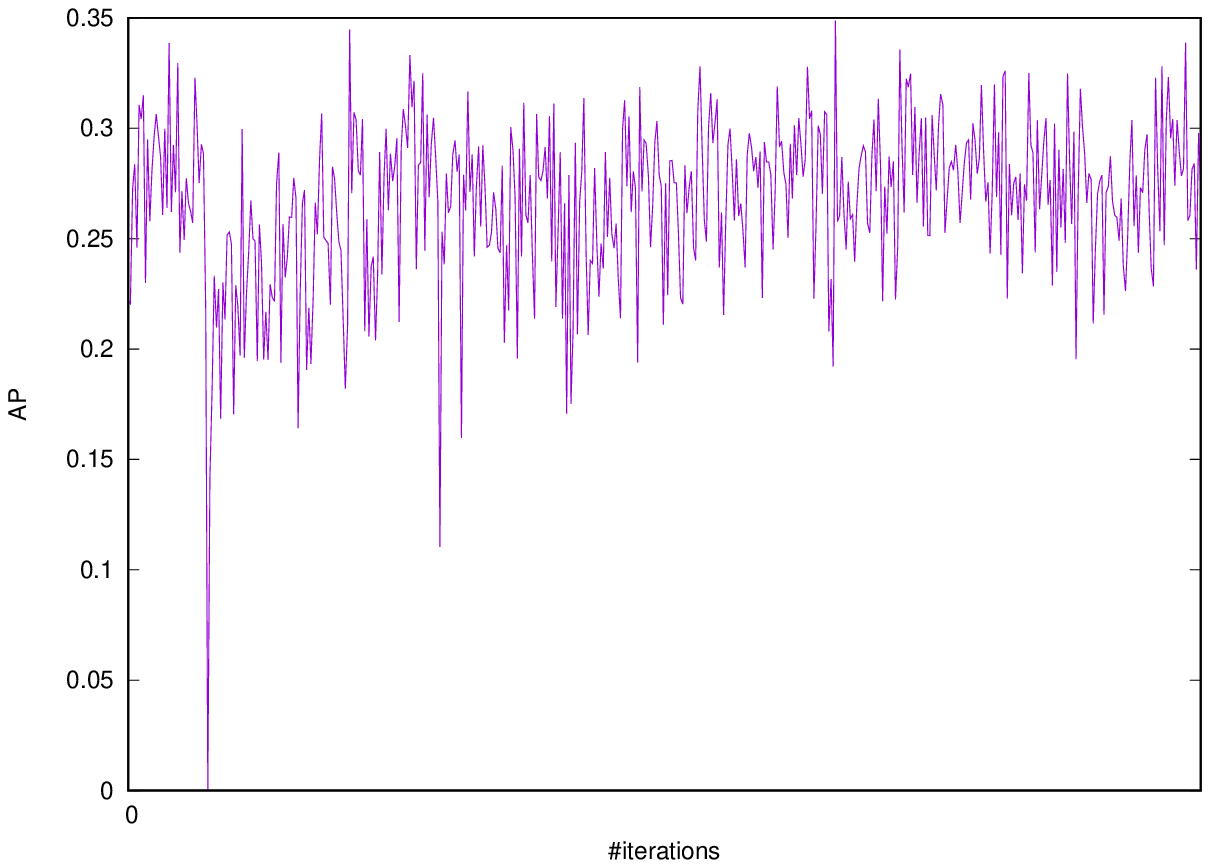}{}\vspace*{8mm}\\
b\\
\FIG{6}{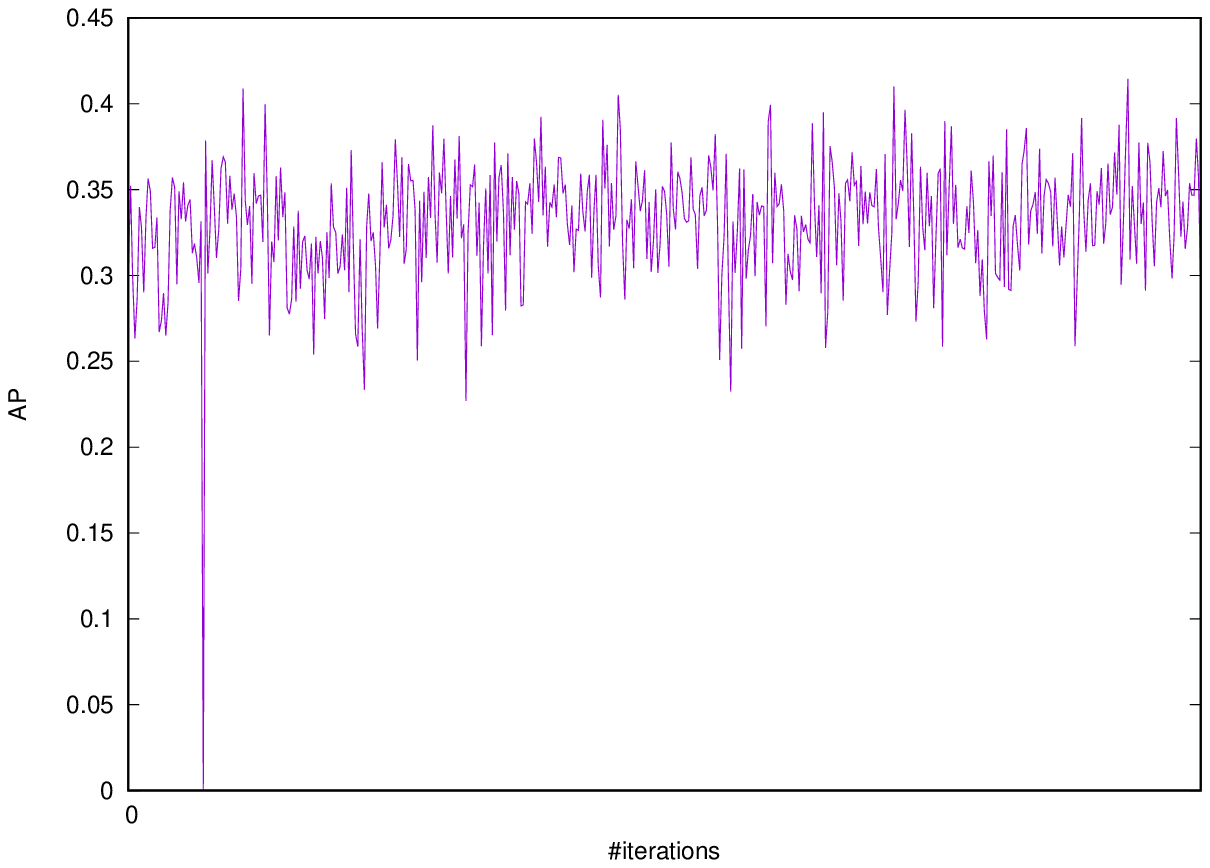}{}\vspace*{8mm}\\
c\\
  \caption{NAS progress.}\label{fig:progress}
\end{figure}
}

\newcommand{\figI}{
  \begin{figure}[t]
    \centering
\FIG{7}{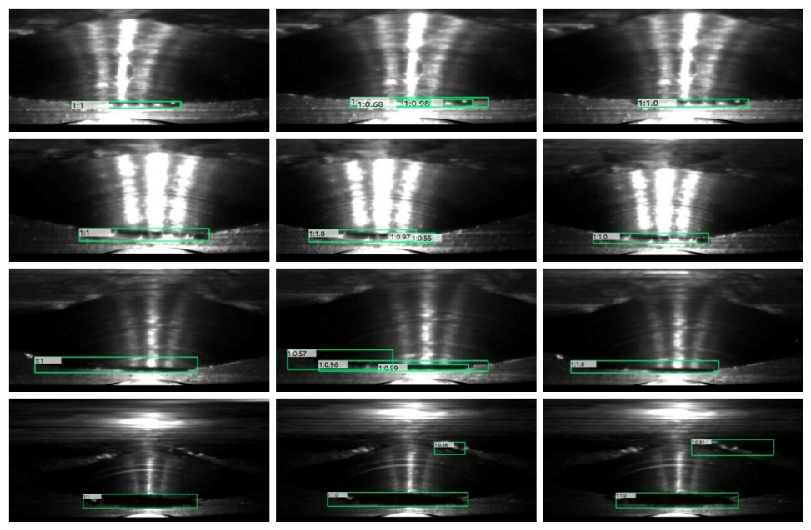}{}
  \caption{VDD results.}
  \label{fig:fig_first}
\end{figure}
}

\title{TaylorMade VDD:\\
Domain-adaptive Visual Defect Detector for High-mix Low-volume Production of Non-convex Cylindrical Metal Objects}

\newcommand{\CO}[1]{}

\author{
Tashiro Kyosuke, Takeda Koji, Tanaka Kanji, Hiroki Tomoe\\
University of FUKUI\\
3-9-1, bunkyo, fukui, fukui, JAPAN\\
{\tt e-mail: \{kyoquick8139, takedakoji00\}@gmail.com, \{tnkknj, hiroki\}@u-fukui.ac.jp}\\
\thanks{We would like to express our sincere gratitude to
Kitagawa Hirofumi,
Takahashi Hisashi,
Matsuno Eduardo,
Takamura Ryota,
and
Takahashi Yasutake,
for development of eye-in-hand system,
which helped us to focus on our VDD project.}
}

\maketitle

\section*{\centering Abstract}
\textit{
Visual defect detection (VDD) for high-mix low-volume production of non-convex metal objects, such as high-pressure cylindrical piping joint parts (VDD-HPPPs), is challenging because subtle difference in domain (e.g., metal objects, imaging device, viewpoints, lighting) significantly affects the specular reflection characteristics of individual metal object types. In this paper, we address this issue by introducing a tailor-made VDD framework that can be automatically adapted to a new domain. Specifically, we formulate this adaptation task as the problem of network architecture search (NAS) on a deep object-detection network, in which the network architecture is searched via reinforcement learning. We demonstrate the effectiveness of the proposed framework using the VDD-HPPPs task as a factory case study. Experimental results show that the proposed method achieved higher burr detection accuracy compared with the baseline method for data with different training/test domains for the non-convex HPPPs, which are particularly affected by domain shifts.
}

\section{Introduction}

In this study, 
we address the problem of 
visual defect detection (VDD) 
for high-mix low-volume production of non-convex metal objects (Fig. \ref{fig:rei1}), such as high-pressure cylindrical piping joint parts (VDD-HPPPs). 
At automatic metal processing site, when drilling holes in metal  using a robot hand, defects called burrs occur. The presence of these burrs often causes scratches and cuts on the hands, which deteriorates safety and affects the accuracy of the product. As HPPPs are produced in small lots, the burr inspection process is not fully automated and demands manual effort. In some factories, the visual inspection is carried out by more than six workers, for 18 hours a day, which is laborious and costly. To address this, automatic eye-in-hand VDD presents a promising solution to this problem.

Automatic VDD on metal objects has been a long standing issue in machine vision literature \cite{classic-vdd} 
and has been energetically studied in various places \cite{czimmermann2020visual,kumar2008computer,xie2008review,huang2015automated,newman1995survey,neogi2014review}. In recent years, there have been attempts to use deep learning for this problem. In \cite{tashiro1}, a flexible multi-layer deep feature extraction method based on CNN via transfer learning was developed to detect anomalies. In \cite{tashiro2}, structural damage detection method based on Faster R-CNN was developed to address the issues of object size variation, overfitting, and specular reflection. In \cite{tashiro3}, remarkable progress was made in detecting corrosion of metal parts such as bolts.

Majority of these existing VDD approaches targeted convex metal objects such as flat steel surface \cite{flat}. Therefore, it is often assumed that the risk of multiple reflections is low and simple foreground/background models were used. In contrast, the HPPPs targeted in this study are non-convex objects. Therefore, subtle difference between training and test domains has a large impact on foreground appearance (i.e., burrs) and background prior. 

\figA
\figB

In this study, we address the above issue by introducing a tailor-made VDD framework (Fig. \ref{fig:eye}) that can be automatically adapted to a new domain (e.g., metal objects, device, viewpoints, lighting). Specifically, we formulate this adaptation task as the problem of network architecture search (NAS) \cite{generic-nas} on a deep object-detection network, in which the network architecture is searched via reinforcement learning. We demonstrate the effectiveness of the proposed framework by using the VDD-HPPPs task as a factory case study. Experimental results show that the proposed method achieved higher burr detection accuracy than the baseline method for data with different domains for the non-convex HPPPs,
which are particularly affected by domain-shift.

\section{Visual Defect Detection (VDD) Problem}

Figure \ref{fig:rei1} shows an example of HPPPs and images taken by 
the eye-in-hand 
in four different domains. 
The arrow indicates the position of the burr.

The HPPPs have a wide variety of registered products, about 20,000 items. The production quantity is mainly small lots, and the monthly production starts from one. This is very different from the conventional applications such as hydraulic parts and engine parts flowing on a dedicated line \cite{non-small-lot}.
Existing image recognition softwares often require as many as seven days to adjust the program to adapt to a new domain, and it is necessary to change the specifications of parts and items.
Therefore, they are not suitable for small lot products such as HPPPs.

In this study, we aim to achieve a good trade-off between the online VDD performance and the offline adaptation speed. To this end, it is necessary to address the following issues. (1) The appearance looks similar between the burr area and the background area in the image (Fig. \ref{fig:rei1}). (2) The shapes and sizes of burrs are diverse and not easy to generalize. (3) The burrs occur in non-convex cylindrical holes inside the joint parts (Fig. \ref{fig:eye}), and are thus affected by the diffused light reflection that is difficult to model. (4) Even when a calibrated eye-in-hand is used, the viewpoint can shift randomly up to about 3 pixels in terms of the image coordinate. To solve the above issues, a highly versatile and accurate machine vision method is needed.

\section{Proposed Approach}

The proposed approach consists of two distinctive stages: the offline-adaptation and online-detection stages. The adaptation stage is responsible for adapting the deep neural network to a new domain, 
and it is the pipeline consisting of 
semi-automatic annotation (\ref{sec:annotation}), 
model-based coordinate-transformation (\ref{sec:transformation}), 
tailor-made network-architecture search (\ref{sec:nas}), 
and network-parameter fine-tuning (\ref{sec:finetuning}). 
The detection stage is responsible for detecting burrs inside a given image, and it consists of 
the coordinate-transformation (\ref{sec:transformation}) and 
visual burr detection (\ref{sec:detection}). Hereinafter, each process will be described in detail.

\subsection{Semi-automatic Annotation}\label{sec:annotation}

Annotation cost is a major part of the total cost required to adapt a VDD software to small-lot metal projects, such as HPPPs.
In our case study's factory site, the annotation is provided by skilled workers in the form of bounding polygons, by using the LabelMe tool \cite{labelme} (Fig. \ref{fig:eye}).
As can be seen from Fig. \ref{fig:rei1}, it would be difficult for a non-skilled person even to visually distinguish burrs from the background textures.
Surprisingly, skilled human workers often become able to distinguish burrs with 100\% accuracy after sufficient time spent training (Fig. \ref{fig:conversion}).
This indicates
that the VDD task may not be infeasible,
which has motivated us
to
develop an automatic VDD system.

\figC

We have been developing a user-interface for semi-automatic annotation in our factory site (Fig. \ref{fig:eye}). 
Using it,
given a 3D CAD HPPP model and a calibrated camera, 
a forward/backward projection model of the camera can be obtained. 
These models enable transferring an annotated bounding polygon 
in the image coordinate of one viewpoint
to 
that of another viewpoint.
This 
transfer technique
eliminates the need for additional annotation 
per different observation condition
on the same HPPP type, 
which leads to significant reduction in the total annotation cost.

\subsection{Model-based Coordinate-transformation}\label{sec:transformation}

Majority of state-of-the-art object detection networks assume bounding box-shaped annotation \cite{jiao2019survey, liu2020deep, zou2019object}.  
In contrast, 
the ground-truth (GT) of 
burr regions in an input image 
often have a 
crescent-like shape
and do not fit well into bounding boxes. To solve this, we propose to transform the image coordinate system appropriately. In our specific case, 
such crescent-like burrs 
usually occur
inside cylindrical HPPPs.
Therefore, 
polar coordinate transformation with the cylinder center coordinates in this image as a hyper-parameter is performed (Fig. \ref{fig:conversion}).
Through preliminary experiments,
we found
that
such a cylinder center
in a given image
can be stably and accurately
predicted
by using a RANSAC-based circle fitting 
to the cylinder border circle.
Figure \ref{fig:conversion} shows the result of image transformation.
Comparing before and after image transformation, it can be confirmed that the filling rate of the burr region with respect to the bounding box is higher after transformation. In the experiment, polar coordinate images with size 
800$\times$1333 pixels were used.

\subsection{Tailor-made Network-architecture Search}\label{sec:nas}

\figD

The proposed NAS framework is inspired by \cite{nas-fpn},
which we 
further developed 
by introducing 
the following two steps, 
which are iterated until the time budget is expired: 
(1) The Controller-RNN creates a candidate architecture and trains a child network with that architecture, and 
(2) The Controller-RNN is updated by policy gradient with rewards obtained from reinforcement learning. 
The objective of the proposed NAS 
is to find an optimal child network,
so as to maximize
the VDD performance 
in terms of
domain adaptation
and
generalization.
In this framework,
a child network is described
by
a parameter variable 
that consists of
a set of 
mutually connected network building blocks
and
their connection relationship (Fig. \ref{fig:D}).
A building block
takes two feature maps from the backbone network (ResNet-50) as input, 
and integrates them using SUM or global pooling operation,
into a new feature map.
Thus,
each block 
can be
described by a triplet ID:
a pair of input feature map IDs
and 
an operation ID ($\in$ \{SUM, POOLING\}).
By definition,
the input IDs chosen are not duplicated.
In addition,
the new 
feature map
created by a block
is regarded
as 
an additional candidate input
for
future building blocks.
Thus,
the space of the triplet ID
can increase
as iteration proceeds.
At each iteration,
the newest feature map is 
considered as the output of a child network.
In the experiments,
a set of 4 
feature maps 
P2, P3, P4 and P5,
with
resolutions 
(200, 334), (100, 167), (50, 84), and (25, 42),
respectively,
are used. 

We observe that
combining 
the
low 
and 
high layer
feature maps 
(i.e., combining primitive and semantic features)
with NAS
is often effective.
The reason might be 
that 
in our application of VDD-HPPPs,
the size of burrs has a large bias and the shape is not constant
and thus,
the high level semantic information plays an important role.
The NAS 
efficiently searches
over the exponentially large number of such combinations,
and
successfully finds
an optimal one,
as we will demonstrate in the experimental section.

The process described in Step-2 is detailed in the following. 
This process aims to maximize the expected reward $J(\theta_c)$ by updating the hyper-parameters $\theta_c$ of the Controller-RNN by policy gradient.
$J(\theta_c)$ is 
defined by:
$J(\theta_c)=E_{P(a_{1:T};\theta_c)}[R]$,
where
$a_{1:T}$ is an action of Controller-RNN. 
Thus, it is updated by:
\[
\bigtriangledown_{\theta_c}J(\theta_c)=\sum_{t=1}^{T}E_{P(a_{1:T};\theta_c)}[\bigtriangledown_{\theta_c}\log{P(a_t|a_{(t-1):1}};\theta_c)R],
\]
where
$a_{1:T}$ is an action of Controller-RNN,
$m=1$ is the number of architectures that the Controller-RNN verifies in one mini-batch.
$T$ is the number of hyper-parameters to be estimated.
$R_k$ is the average-precision (AP) in the $k$-th architecture.
$b$ is the exponential moving average of the AP of the neural network architecture up to that point.
The smoothing constant is 0.8.
The learning rate is 0.1.
The number of training iterations is 3000 per child network.
The hyper-parameters in \cite{nas-fpn}
are set as follows:
cfgs.LR = 0.001,
cfgs.WARM-step = 750, 
and
\#trials = 500.
The NAS with above setting 
consumes 
2 weeks
using a graphics processing unit (GPU) machine (NVIDIA RTX 2080).
Figure \ref{fig:progress} 
shows 
progress of NAS search.

\subsection{Network-parameter Fine-tuning}\label{sec:finetuning}

The searched architecture is then used to train a burr detector. For the training, the Faster R-CNN algorithm is used. 
At this time, the parameters for training are set as followings. cfgs.LR=0.001. cfgs.WARM-step=2,500. Subsequent learning rates are reduced 1/10 times when the number of training sessions is 60,000, and 1/100 times when the number of training sessions is 80,000. The number of training iterations is 150,000. The time required is about 2 days using the above same GPU machine.

\subsection{Visual Burr Detection}\label{sec:detection}

The detection process takes a query image and predicts bounding boxes of burr regions with a non-maxima suppression. The confidence score is evaluated as the highest probability values among all the classes in the Faster R-CNN. The computation speed per image was around 3 fps.

\section{Experiment}

The proposed tailor-made VDD framework has been evaluated 
using real HPPP dataset 
in four different domains
that is collected on the target factory site.
This section
describes
the dataset, the baseline method, experimental results,
and provides discussions.

\subsection{Dataset}

Figure \ref{fig:rei1} shows examples of image datasets. 
We collected four independent collections of images, A, B, C, and D at different domains, and manually annotated every image.
The dataset size are 402, 396, 50, and 76 for A, B, C, and D, respectively.
The set A is used for NAS, training and testing,
while B, C and D are used only for testing. 

The set A is split into 1:2:1 subsets 
namely, 
NAS, training, and evaluation subsets. 
The NAS subset is used for evaluating each child network during the NAS task. 
The union of the NAS and training subsets is used for training.  The evaluation subset is used 
for performance evaluation 
on a trained NAS-searched VDD model.  
As aforementioned, every training/test image is transformed to polar coordinate before being input to the training or testing procedure. 
For NAS and training, a left-right flipping data augmentation is applied.

\subsection{Baseline Method}

A deep object detector using a feature pyramid network (FPN) \cite{fpn} is used as a baseline method. The FPN consists of three features, bottom-up direction, top-down direction, and potential connection, to the feature layers of different scales output by the convolutional neural network. 
This provides 
both low-resolution semantically strong features 
and 
high-resolution semantically weak features. 
A Faster R-CNN is used for detection task.
Despite the efficiency, FPN is based on a manually designed architecture, 
and thus,
it is not optimized 
for
a given specific application.

\subsection{Results}

Figure \ref{fig:progress}
shows
NAS progress
when
subset A is used as the NAS subset.
It was confirmed 
that the 
curve rises slightly as the trial proceeds,
and that
the Controller-RNN learned the generation of a better feature map over time through trial and error, and the search was performed adequately.

\figH

\tabAB

As shown in the Fig. \ref{fig:progress}, 
the NAS score converged at 110,000-th iteration. We use the architecture at this point to evaluate the VDD performance. Specifically, the convergence is judged if change in AP values between two consecutive training sessions is equal or lower than 0.01. 
For performance evaluation,
average precision (AP) is evaluated 
for different threshold values on IoU,
from 0.5 to 0.95 with 0.05 increment step,
and then
average of these AP values 
is used as the performance index.

Table \ref{tab:AB}
shows performance results.

Let us discuss the results for 
trained NAS searched VDD model
using the dataset A as the training set.
From the evaluation result,
it is confirmed that 
the proposed method
provides better 
performance
compared with FPN 
in all the test sets.
Figure \ref{fig:fig_first}
shows
example detection results
for
GT/FPN/Proposed (columns)
for 
test domains A/B/C/D (rows).
with bounding boxes
whose
confidence scores
are higher than 0.5.
Overall,
the number of bounding boxes 
generated by the proposed method 
tended to be the same
as that of the ground-truth.
Exceptionally,
for testing domain D,
there were
much
false positives 
for both the proposed and FPN methods.
This is mainly due to the dark lighting conditions.
It should be noted 
that
for FPN, 
the bounding boxes often do not appear,
as the confidence score
is often lower than 0.5.

From the above results, it could be concluded that 
VDD performance was significantly improved 
by the proposed taylor-made VDD 
despite the fact
that the adaptation process is highly automated and efficient.

\figI

\section{Conclusion}

In this study, we presented a tailor-made visual defect detection framework that can be adapted to various domains. To the best of our knowledge, we are the first 
to formulate the VDD-HPPPs as an important and challenging new 
machine-vision application that is characterized 
by 
a combination of
non-convex metal parts with complex specular reflections and high-mix low-volume production. In this study, we adapted the NAS technique to search for the 
optimal architecture of the network. Through a factory case study, we demonstrated that our approach is able to search for a versatile network architecture and enables us to detect burrs with higher accuracy compared with the baseline method under domain shifts.

\bibliographystyle{IEEEtran}
\bibliography{amc}

\end{document}